\documentclass{article}
\usepackage{authblk}

\usepackage{graphicx}
\usepackage{booktabs,multirow}
\newcommand{\first}[1]{\textbf{#1}}

\usepackage[accsupp]{axessibility}  

\usepackage{algorithm}
\usepackage{algorithmic}

\newcommand{\HuC}{HuC-VideoMAE}
\newcommand{\CompleteHuC}{Human Centric Video Masked Auto Encoder}

\usepackage{orcidlink}
\usepackage{hyperref}

\begin{document}

\title{\HuC: Human-Centric Video Masked Autoencoding from synthetic data}

\author[1]{Ricardo Pizarro\orcidlink{0000-0001-7894-0236}, }
\author[2]{Roberto Valle\orcidlink{0000-0003-1423-1478}}
\author[3]{Jos\'e M. Buenaposada\orcidlink{0000-0002-4308-9653}}
\author[1]{Luis M. Bergasa\orcidlink{0000-0002-0087-3077}}
\author[2]{Luis Baumela\orcidlink{0000-0001-6910-4359}}
\affil[1]{Universidad de Alcal\'a, Alcal\'a de Henares, Spain}
\affil[2]{Universidad Polit\'ecnica de Madrid, Madrid, Spain}
\affil[3]{Universidad Rey Juan Carlos, M\'ostoles, Spain}
\date{ricardo.pizarroc@edu.uah.es}

\maketitle

\begin{abstract}
Modern action recognition models rely on video transformers pretrained on massive collections of web-crawled videos, such as Kinetics-700. However, the use of such data raises ethical concerns, as subjects' consent is typically not obtained. Recent high-quality synthetic video datasets generated from motion-capture data, such as BEDLAM2.0, offer a promising ethical alternative.
In this work, we investigate self-supervised pretraining of video transformers on synthetic human-motion datasets. We first show that directly applying the standard VideoMAE masking strategy leads to substantially worse performance than pretraining on Kinetics. To address this limitation, we propose a human-centric masking scheme that leverages body keypoints and person bounding box regions. Our approach encourages the model to focus on the structure and dynamics of human motion during pretraining.
Experiments on NTU RGB+D and Toyota-Smarthome demonstrate that our method significantly outperforms standard VideoMAE pretraining on synthetic data, closing $49\%$ of the gap to Kinetics pretraining on NTU RGB+D cross-view-subject without using a single real frame during pretraining. To promote the use of ethical action recognition models, we will publicly release our pretrained models upon acceptance. Code and trained models are available at \url{https://github.com/pcr-upm/eccvw26_recognition}
\end{abstract}

\section{Introduction}
\label{sec:intro}

Modern human motion video understanding is built on transfer learning where a video transformer~\cite{tong2022videomae,liu2022videoswing,li2022mvitv2} is first pretrained on a massive corpus of short clips and then fine-tuned on a downstream task~\cite{reilly2024just,poguise2025}. The pretraining corpora that make this work, e.g. Kinetics~\cite{kay2017kinetics}, are assembled by scraping web videos at scale, capturing millions of identifiable people who never consented to their likeness, gait or behaviour being used to train models. This conflicts with emerging data-protection norms such as the GDPR~\cite{gdpr2016} and the EU AI Act~\cite{euaiact2024}. The problem is that video information is intrinsically biometric, since a person's identity can be recovered from appearance~\cite{ye2022reid} or gait~\cite{sepasmoghaddam2023gait} even when a dataset is ``de-identified''. As action-recognition systems are deployed in privacy-sensitive settings such as, smart homes~\cite{das2019toyota}, or human-robot interaction~\cite{synthact}, the provenance of the pretraining data becomes an ethical liability that fine-tuning on a clean downstream set does not erase.

Computer-graphics rendering of motion-captured humans offers an alternative. Datasets such as SURREAL~\cite{varol2017surreal}, the AMARV corpus~\cite{synthact} and, most recently, BEDLAM2.0~\cite{tesch2025bedlam2} generate photorealistic clothed avatars driven by consented motion-capture archives, with full control over appearance, viewpoint and scene. No real person is depicted, so the pretraining data carries no consent or biometric-leakage risk by construction, while remaining large and diverse enough for self-supervised learning. This raises an interesting question for privacy-preserving vision, can we pretrain video transformers entirely on ethically-sourced synthetic humans without sacrificing downstream accuracy?

Answering it is not simply a matter of swapping datasets. We find that taking the popular self-supervised recipe VideoMAE~\cite{tong2022videomae}, which reconstructs clips under high-ratio random ``tube'' masking, and applying it unchanged to synthetic video yields representations that transfer substantially worse than the same model pretrained on Kinetics. We argue this gap is a property of the objective, not only of the data. The reason is that uniform masking is agnostic to what it hides and at a $90\%$ masking ratio the reconstruction budget is dominated by background and rendered texture, and the human is left almost entirely visible or entirely hidden by chance. On public real web video data sets, this is a slight inefficiency; although there is enough scale, diversity, and data to overcome this. Meanwhile, on the available public synthetic data sets, with a much lower amount of video clips and where backgrounds and textures are comparatively easy to learn, it allows the model to shortcut the task without learning human motion, widening the synthetic-to-real gap.

Our experiments show that synthetic data already hands us, for free, the structural supervision needed to alleviate this. Every rendered frame comes with ground-truth $2$D body keypoints, a person bounding box and a background separation, signals that on real video would require noisy off-the-shelf estimation~\cite{huang2023mgmae}. We use these signals to make masking human-centric where each patch receives a masking score from its semantic region. The background is masked most aggressively, the actor's body is masked nearly as heavily, and only the sparse keypoints are kept most visible, acting as structural anchors for the reconstruction targets. The keypoints are masked at the identity level (e.g. right wrist or left elbow) so that a joint chosen for masking stays hidden across the whole clip rather than reappearing in a neighboring frame. This concentrates the reconstruction signal on the body and its motion. We call the resulting method \emph{\CompleteHuC}~(or \HuC{} for short).

\HuC{} partially compensates for the degradation introduced by random tube masking. On NTU RGB+D, it increases the cross-view-subject average accuracy from $80.1\%$ to $85.8\%$, reducing by $49\%$ the gap between the baseline and a Kinetics-pretrained model ($91.7\%$ average accuracy). Notably, these gains are obtained without access to real frames during pretraining. The method also maintains consistent performance across viewpoints, with a decrease of $1.8$ percentage points from $0^\circ$ to $90^\circ$. On Toyota-Smarthome, \HuC{} achieves higher accuracy than a Kinetics-pretrained VideoMAEv2 baseline on CV1 and performs within $4$ percentage points of that baseline on CV2.

Our contributions are:
\begin{itemize}
    \item We investigate self-supervised pretraining on synthetic motion-capture humans as a privacy-preserving alternative to web-crawled video. We show that the standard VideoMAE masking objective transfers less effectively in this setting, highlighting the need for masking strategies tailored to synthetic human data.
    \item We propose \HuC{}, a human-centric masking strategy that leverages the ground-truth keypoints, person masks, and background information available in synthetic data sets. The method introduces temporally consistent, identity-level keypoint masking and is evaluated on both ViT and MViTv2 backbones. 
    \item We demonstrate that \HuC{} improves transfer performance on NTU RGB+D and Toyota-Smarthome while maintaining stable performance across viewpoints. 
\end{itemize}

\section{Related work}

\paragraph{Masked video modelling.}
Works like BEiT~\cite{bao2022beit} and the Masked Autoencoder (MAE)~\cite{he2022mae} showed that reconstructing heavily masked inputs yields strong transferable representations. VideoMAE~\cite{tong2022videomae} extended this to video with tube masking along the temporal dimension at very high ratios ($90\%$), and VideoMAEv2~\cite{wang2023videomaev2} scaled it with dual masking, while MaskFeat~\cite{wei2022maskfeat} reconstructs HOG features instead of pixels. 
An issue with uniform masking probability is that regions occupying a larger fraction of the image, such as the background, contribute disproportionately to the reconstruction objective. To address this, several works have introduced content-aware masking strategies. MGMAE~\cite{huang2023mgmae} warps masks along estimated optical flow to keep them on moving regions. SMILE~\cite{thoker2025smile} overlays synthetic moving objects and masks along their trajectories while reconstructing CLIP features. MARLIN~\cite{marlin2023} masks tubes guided by semantic facial regions (eyes, nose, mouth, skin) to learn transferable facial representations.

While these approaches improve the allocation of masking across informative regions, they largely retain the conventional pretraining paradigm based on large-scale collections of web-sourced video. Consequently, although self-supervision reduces dependence on manual annotations, it does not address concerns related to the origin, consent, or privacy characteristics of the underlying video data. Meanwhile, synthetic sources provide ground-truth data that can guide the informative regions without the need to rely on noisy estimates such as optical flow and facial segmentation.

\paragraph{Synthetic data and ethical pretraining for action recognition.}
Rendering humans from motion capture is an established way to obtain labeled video without collecting real footage. Early pipelines such as PHAV~\cite{desouza2017phav} SURREAL~\cite{varol2017surreal}, and SURREACT~\cite{varol2021synthetic} showed that synthetic humans rendered from novel viewpoints improve cross-view action recognition on NTU RGB+D~\cite{shahroudy2016ntu,liu2019ntu120}. More recently, SynthAct~\cite{synthact} introduced the large AMARV corpus for synthetic-to-real pretraining, and BEDLAM2.0~\cite{tesch2025bedlam2}, originally built for $3$D human pose and shape estimation, provides highly realistic clothed humans in diverse scenes. 

These datasets are appealing as an ethical alternative to web-crawled corpora such as Kinetics~\cite{kay2017kinetics}, whose subjects have not consented to their use. Prior work, however, either trains supervised classifiers on the synthetic labels or applies off-the-shelf masking. We instead ask how the \emph{self-supervised objective} should be designed to make ethically-sourced synthetic pretraining competitive, and show that a human-centric masking scheme substantially narrows the gap to real-data pretraining.

\section{Human centric video masked autoencoder pretraining}
\begin{figure*}
  \centering
  \includegraphics[width=\textwidth]{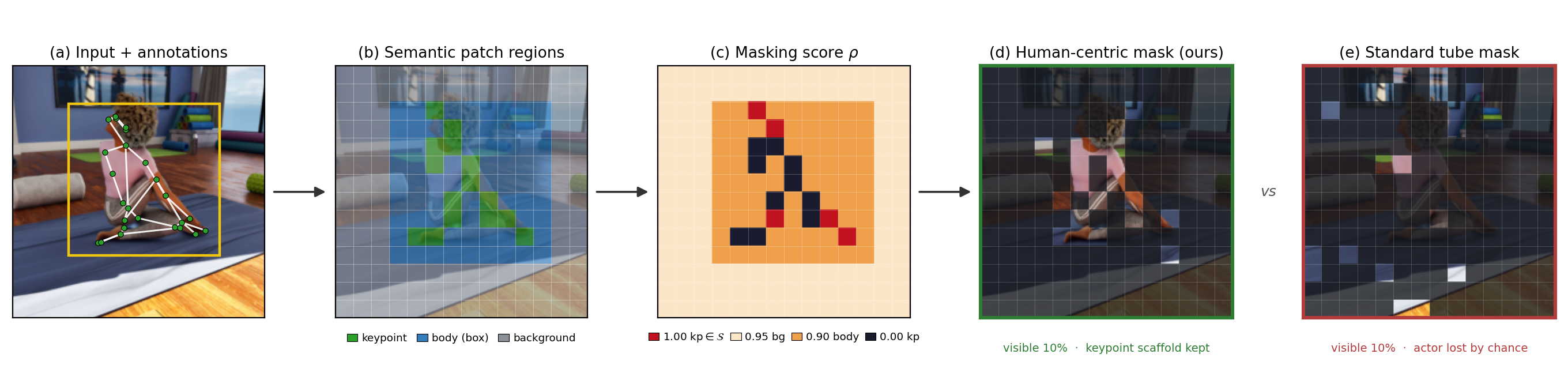}
  \caption{\textbf{Body-keypoint-guided masking.} From a synthetic frame with $2$D keypoint and person-box annotations~(a), every patch is labelled as \emph{keypoint}, \emph{body} (inside the box) or \emph{background}~(b), and receives a masking score $\rho$~(c): background scores above the body, and keypoints lowest of all, so that the sparse joint structure survives the mask. The exception are the joints selected for identity masking ($\in\mathcal{S}$, red), with score above the background and are therefore always masked.}
  \label{fig:masking}
\end{figure*}

We build on the VideoMAE~\cite{tong2022videomae,wang2023videomaev2} masked autoencoding framework, in which a video transformer is pretrained to reconstruct a large fraction of masked spatio-temporal patches from a small set of visible ones. We use the VideoMAEv2~\cite{wang2023videomaev2} implementation with the original single-mask objective (the decoder reconstructs all masked positions) and ViT backbones together with the MViTv2~\cite{li2022mvitv2} backbones from PyTorch. Each clip is split into patches and neighbouring frames are grouped into tubelets, yielding a spatio-temporal token grid on which masking operates. Our contribution is a new \emph{masking policy}. Rather than masking patches uniformly at random, we exploit the structural annotations that synthetic data provides, per-frame $2$D body keypoints and person bounding boxes, to concentrate the reconstruction signal on the human actor.

\subsection{Body-keypoint-guided masking}
\label{subsec:kp_masking}

\begin{figure}
  \centering
  \includegraphics[width=\linewidth]{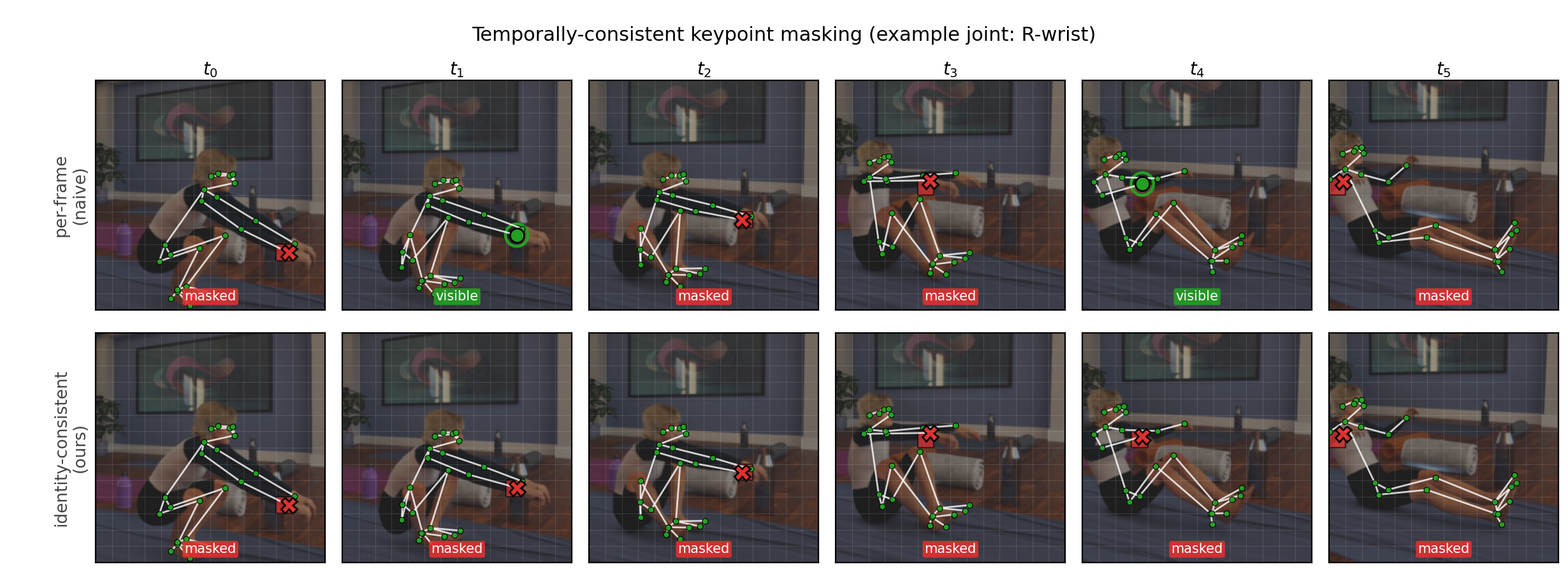}
  \caption{\textbf{Keypoint masking is temporally consistent.} A keypoint selected for masking (here the right wrist) is hidden in \emph{every} frame of the clip (bottom row), because the choice is made at the body part identity level. A naive per-frame choice (top row) would leave the same joint visible in some frames ($t_1,t_4$), letting the decoder copy it from a neighbouring frame. Green: visible keypoints; red: the masked joint and its patch.}
  \label{fig:kptemporal}
\end{figure}

Standard tube masking treats every patch equally, so at a high masking ratio most of the reconstruction budget is spent on background regions that are irrelevant to action recognition, and the actor can be almost entirely visible or almost entirely hidden by chance. We instead assign every patch a \emph{masking score} derived from its semantic region, and mask the highest-scoring patches first (Fig.~\ref{fig:masking}). Each synthetic frame is annotated with the body keypoints and a person bounding box, obtained directly from the BEDLAM2.0 labels. After the random crop-and-resize augmentation, keypoints and boxes are mapped through the same crop parameters into the patch grid, so the mask stays spatially aligned with the augmented input. Each patch then falls into one of three regions with a distinct score:

\begin{itemize}
    \item \textit{background} (outside the person box): highest score $\rho_{bg}$, i.e. masked first,
    \item \textit{body} (inside the bounding box but not on a joint): intermediate score $\rho_{body}$,
    \item \textit{keypoint} (containing a projected joint): lowest score$=0$, if kept visible, or score$=1$, if its keypoint identity is selected as masked ($\in\mathcal{S}$).
\end{itemize}
with $\rho_{bg} > \rho_{body} > 0$ and not-masked keypoint patches scoring below both. The background is masked almost completely, and the actor's body is masked nearly as aggressively, so most of the reconstruction budget falls on these two. Keypoints are kept mostly visible, they act as sparse structural anchors rather than reconstruction targets, giving the decoder pose cues from which to infer the motion of the surrounding masked body patches. A fraction $\rho_{kp}$ of the \emph{keypoint identities} (e.g. left wrist, right ankle, right wrist) is selected once per clip and forced to be masked in every frame. Since the selection is made at the identity level and re-used across temporal slots (tubelets), the same joints are hidden throughout the clip (Fig.~\ref{fig:kptemporal}), which prevents the decoder from trivially copying a joint that is visible in an adjacent frame.
As in standard tube masking, the number of masked patches per frame is kept fixed, ensuring a constant number of visible tokens across the batch. Specifically, for a frame containing $N_{\mathrm{p}}$ patches, we mask $K = r\cdot N_{\mathrm{p}}$, where $r$ is the target masking ratio. Each patch is assigned a region-dependent masking score, and the $K$ highest-scoring patches are selected for masking. The complete masking procedure is summarized in Algorithm~\ref{alg:huc}.

\begin{algorithm}[t] 
\caption{\HuC{} Mask Generation} 
\label{alg:huc} 
\begin{algorithmic}[1] 
\STATE $\mathcal{T}\gets$ body keypoint identities visible in at least one frame \STATE $m \gets \lfloor \rho_{kp} \cdot |\mathcal{T}| \rfloor$ \STATE $\mathcal{S}\gets\mathrm{RandomSubset}(\mathcal{T}, m)$
\STATE $\mathrm{jitter}(p)\gets\mathcal{U}(0,10^{-3})$ for every patch $p$, drawn once per clip
\STATE $K \gets \lfloor r\cdot N_{\text{p}} \rfloor$
\FOR{each  temporal slot $t$} 
    \STATE Project keypoints and person bounding box, $B_t$, to the patch grid 
        \FOR{each patch $p$}             
            \IF{$p$ contains a keypoint whose identity is in $\mathcal{S}$}
                \STATE $\mathbf{s}(p)\gets 1$ \COMMENT{$>\rho_{bg}$, always masked}
            \ELSIF{$p$ contains a keypoint}
                \STATE $\mathbf{s}(p)\gets 0$ \COMMENT{lowest, kept visible}
            \ELSIF{$p \in B_t$}
                \STATE $\mathbf{s}(p)\gets \rho_{body}$
            \ELSE
                \STATE $\mathbf{s}(p)\gets \rho_{bg}$
            \ENDIF
        \ENDFOR 
        \STATE $\mathbf{s} \gets \mathbf{s} + \mathrm{jitter}$ \COMMENT{randomizes selections within a region}
        \STATE Rank all patches by score $\mathbf{s}(p)$ 
        \STATE $\mathcal{M}_t \gets$ indices of the top-$K$ scoring patches 
        \STATE Mask all patches in $\mathcal{M}_t$
\ENDFOR
\end{algorithmic} 
\end{algorithm}

Naturally, this policy forces the model to reconstruct the body region from the sparse structure of visible joints and their motion, rather than exploiting background regularities, encouraging actor-centric and viewpoint-robust representations.

\section{Experiments}

In this section, we evaluate the impact of our human-centric masking strategy on self-supervised video pretraining. Models are first pretrained on the synthetic BEDLAM2.0 dataset~\cite{tesch2025bedlam2} and then fine-tuned on downstream action recognition benchmarks, namely NTU RGB+D 60 and Toyota-Smarthome. To contextualize the results, we compare against VideoMAEv2 models pretrained on Kinetics~\cite{kay2017kinetics} as well as the AMARV synthetic pretraining approach~\cite{synthact}. All methods are evaluated under the same fine-tuning and testing protocols.

\subsection{Datasets}
\label{subsec:datasets}

BEDLAM2.0~\cite{tesch2025bedlam2} is a large-scale synthetic human-motion dataset generated from motion-capture sequences rendered with realistic human body models in diverse virtual environments. The dataset features realistic camera configurations, including varying focal lengths and dynamic motions such as panning, zooming, orbiting, and tracking. It contains more than 27K image sequences, over 8M frames, and more than 4K distinct body shapes, resulting in approximately 13.3M person annotations. In addition to RGB video, BEDLAM2.0 provides accurate human-centric supervision, including 3D poses, projected 2D keypoints, and person bounding boxes. The combination of large scale, rich annotations, and synthetic data generation makes it an attractive benchmark for studying privacy-preserving, human-centric self-supervised video pretraining. We use BEDLAM2.0 exclusively during pretraining.

The AMARV (Archive of Motion Capture As Rendered Videos) dataset~\cite{synthact} is a large-scale, synthetic data set specifically optimized for training generalizable human action recognition and computer vision models. Generated via the automated SynthAct pipeline, AMARV provides over $0.8M$ single-person multi-view synthetic video clips derived from extensive underlying human motion capture databases. Each action sequence is simultaneously rendered from four synchronized, static viewpoints, generating both highly detailed RGB footage and corresponding depth maps. Alongside the video data, the dataset delivers a comprehensive suite of frame-level and clip-level ground truth annotations, including exact 3D body poses, surface normal maps, semantic segmentation masks, and bounding boxes.

Compared to AMARV, BEDLAM2.0 emphasizes diversity and realism over clip count. Although AMARV contains more action clips, BEDLAM2.0 includes over 4K body shapes (compared to 106 in AMARV), a wider variety of clothing and accessories, and more diverse scene layouts. Furthermore, BEDLAM2.0 models realistic camera behavior through motions such as panning, zooming, orbiting, and tracking, whereas AMARV uses static camera viewpoints. As a result, AMARV primarily provides large-scale multi-view action observations, while BEDLAM2.0 exposes the model to greater variability in human appearance, scene context, and camera dynamics.

Original VideoMAEv2 is pretrained first in a 1.35M set of videos and then fine-tuned in a supervised way in the union of Kinetics 400, 600 and 700 (i.e. the number of action categories) with 710 categories and 0.66M clips. BEDLAM2.0 contains approximately 27K clips, compared to the 1.35M clips used for VideoMAEv2 pretraining, representing a roughly $50\times$ smaller corpus (about 1.7 orders of magnitude fewer clips).

NTU RGB+D 60~\cite{shahroudy2016ntu} is a widely used benchmark for human action recognition in indoor environments. The dataset contains approximately 57K videos covering 60 action classes performed by 40 subjects and recorded from multiple camera viewpoints. We follow the cross-view-subject protocol introduced by SURREACT~\cite{varol2021synthetic}, which extends the standard cross-subject split, where training and test subjects are disjoint, by additionally restricting training to front-view videos only, and then testing separately on the $0^\circ$, $45^\circ$ and $90^\circ$ views. Accuracy at $45^\circ$ and $90^\circ$ therefore measures generalization to viewpoints never seen during fine-tuning Following common practice, we report overall classification accuracy (\emph{Acc.}). 

Toyota-Smarthome~\cite{das2019toyota} focuses on activities of daily living performed in a realistic home environment. The dataset contains around 16K RGB clips from 31 activity classes performed by 18 subjects and captured from seven camera viewpoints. We evaluate using the cross-view protocols CV1 and CV2, which use a 19-class subset. In CV1, training uses only camera 1, while camera 2 is reserved for testing. In CV2, training uses cameras 1, 3, 4, 6, and 7, while camera 5 is used for testing. Due to class imbalance, we report mean class accuracy (\emph{mCA}).

\subsection{Implementation details}
\label{subsec:impl}
We sample clips of $16$ frames at a temporal stride of $4$, resized to $224\times224$; frames are split into $16\times16$ patches with tubelet size $2$, giving an $8\times14\times14$ token grid. We use ViT-B and MViTv2-S backbones with a $4$-layer reconstruction decoder and single-mask decoding (decoder mask ratio $0$). For the baseline pretrained models we use the weights \texttt{vit\_b\_k710\_dl\_from\_giant} and \texttt{pretrain\_mvitv2\_small\_224}. We mask $90\%$ ($r=0.9$) of the patches per temporal slot, with region scores $\rho_{bg}=0.95$ and $\rho_{body}=0.90$, and mask a fraction $\rho_{kp}=0.2$ of the keypoint identities per clip. Pretraining runs for $200$ epochs with AdamW ($\beta=(0.9,0.95)$), a base learning rate of $10^{-3}$, a $20$-epoch warm-up with cosine decay, and $4$ augmented views per clip.
For MViTv2-S, the mask is first computed on the token grid and then expanded to the corresponding $2\times32\times32$ spatio-temporal macro-patch in pixel space. This preserves the regular grid structure required by MViTv2's hierarchical pooling attention while keeping both the masking ratio and the region-dependent masking preferences unchanged.

\subsection{Training and Evaluation Procedure}
\label{subsec:training_eval_procedure}

To evaluate the transferability and cross-view generalization of our models, we adopt a two-stage paradigm consisting of self-supervised pretraining followed by supervised downstream fine-tuning.

\paragraph{Pretraining stage.}
We compare three initializations of the VideoMAE (ViT-B) backbone:
\begin{itemize}
    \item \textit{VideoMAE (Kinetics)}: the publicly released weights self-supervised on the real-world Kinetics dataset with random tube masking, serving as an upper reference for web-crawled pretraining.
    \item \textit{VideoMAE (BEDLAM2.0)}: the backbone pretrained from scratch on the synthetic BEDLAM2.0~\cite{tesch2025bedlam2} dataset with the same random tube masking, isolating the effect of switching the pretraining source from real to synthetic data.
    \item \textit{\HuC{}}: pretrained on BEDLAM2.0 with our proposal of body-keypoint-guided masking presented in section \ref{subsec:kp_masking}.
\end{itemize}
Both synthetic variants share identical data, schedule and hyper-parameters, so any downstream difference is attributable to the masking policy alone.

\subsection{Effect of the masking policy}
\label{subsec:ablation}

In Table~\ref{tab:ablation} the backbone, optimizer, schedule and augmentation are held fixed. Only the pretraining source (real Kinetics vs.\ synthetic BEDLAM2.0) and the masking scheme (random tube masking vs.\ our body-keypoint-guided \HuC{}) change, so any difference is attributable to that one factor.

\begin{table}
\centering
\small
\setlength{\tabcolsep}{8pt}
\caption{\textbf{Ablation of the masking policy on NTU RGB+D, cross-view-subject (ViT-B).} The backbone, optimizer, schedule and augmentation are identical across rows. Only the pretraining source and the masking scheme change. All numbers are downstream fine-tuning accuracy (in $\%$) after synthetic-to-real transfer (S$\rightarrow$R), except the Kinetics row (R$\rightarrow$R).}
\label{tab:ablation}
\begin{tabular}{l|l|c|c|c|c}
\toprule
Pretrain data & Masking & $0^\circ$ & $45^\circ$ & $90^\circ$ & Avg \\
\midrule
Kinetics & random tube & 92.2 & 91.8 & 91.2 & 91.7 \\
\hline
BEDLAM2.0  & random tube & 81.6 & 79.6 & 79.1 & 80.1 \\
BEDLAM2.0 & \HuC{} & 86.6 & 85.9 & 84.8 & \first{85.8} \\
\bottomrule
\end{tabular}
\end{table}

On NTU RGB+D cross-view-subject (ViT-B backbone), switching the pretraining source from Kinetics to BEDLAM2.0 while keeping random tube masking costs $11.6$ points of accuracy averaged over the three viewpoints ($91.7\rightarrow80.1$). This is the synthetic-to-real gap caused by naively applying random masking, at $90\%$ masking ratio, most of what the model is asked to reconstruct is easily-learned background and rendered texture, so it has little incentive to model the human actors. Replacing random masking with \HuC{} on the same synthetic data and schedule recovers $5.7$ of these $11.6$ points ($80.1\rightarrow85.8$), closing $49\%$ of the gap to real-data pretraining without using a single real frame in pretraining.

\subsection{Effect of the synthetic corpus}
\label{sec:synthcorp}

An important question is how much the choice of synthetic corpus itself matters, independently of the masking policy. Table~\ref{tab:source_ablation} compares VideoMAE pretrained with uniform tube masking on BEDLAM2.0~\cite{tesch2025bedlam2} against the same recipe pretrained on the AMARV corpus~\cite{synthact} and fine-tuned on downstream datasets. BEDLAM2.0 outperforms AMARV by a wide margin on both NTU RGB+D ($0^\circ/45^\circ/90^\circ$) at $81.6/79.6/79.1$ for BEDLAM2.0 versus $57.94/53.59/53.56$ for AMARV. With $80.1$ vs.\ $55.0$ average across viewpoints ($+25.1$), and Toyota-Smarthome  $54.1/58.86$ for BEDLAM2.0 versus $41.71/42.87$ ($56.5$ vs.\ $42.3$ average across CV1/CV2, $+14.2$), confirming that BEDLAM2.0 is a stronger pretraining dataset regardless of the masking policy.

\begin{table}
\centering
\small
\setlength{\tabcolsep}{8pt}
\caption{\textbf{Strength of the pretraining source, independent of masking policy.} Random tube masking, same backbone and schedule. Only the synthetic pretraining corpus changes. Downstream accuracy averaged per dataset (NTU: $0^\circ/45^\circ/90^\circ$. Toyota: CV1/CV2).}
\label{tab:source_ablation}
\begin{tabular}{l|c|c}
\toprule
Pretrain data & NTU (Avg) & Toyota (Avg) \\
\midrule
AMARV (synth.) & 55.0 & 42.3 \\
BEDLAM2.0 (synth.) & \first{80.1} & \first{56.5} \\
\bottomrule
\end{tabular}
\end{table}

The comparison between AMARV and BEDLAM2.0 shows that raw scale alone does not explain the gap. AMARV is the larger corpus by clip count ($800$K multi-view clips vs.\ BEDLAM2.0's 
$27K$ video sequences) and draws on more distinct source motions ($10{,}892$ AMASS/BABEL sequences vs.\ $4{,}643$ for BEDLAM2.0)~\cite{synthact,tesch2025bedlam2}, yet it is BEDLAM2.0 that transfers far better.  (Table~\ref{tab:source_ablation}).
We attribute this to more body diversity in BEDLAM2.0 (4K body shapes vs 106), realistic camera motion and rendering quality. BEDLAM2.0 uses simulated clothing, strand-based hair, modelled $3$D environments and dynamic camera motion, whereas AMARV maps static, unsimulated clothing onto bodies in front of flat HDRI backdrops with cameras fixed per clip.

\subsection{Comparison with the state of the art}
\label{subsec:sota}

In this section, we perform supervised fine tuning of the previously trained VideoMAE models with self-supervision. We use the NTU RGB+D and then the Toyota-Smarthome as representatives of the action recognition datasets used in the literature.

\begin{table}[h]
\centering
\setlength{\tabcolsep}{8pt}
\caption{Cross-view-subject evaluation on NTU RGB+D, comparison with SURREACT\cite{varol2021synthetic} and our new VideoMAE-based results. S $\rightarrow$ R, Pretrained on Synth and then fine-tuned on Real.
}
\label{tab:cross_view_subject}
\begin{tabular}{lr|c|ccc}
\toprule
\multicolumn{2}{l|}{Model} & Pretrain Data & $0^\circ$ & $45^\circ$ & $90^\circ$ \\ \hline
SURREACT~\cite{varol2021synthetic}  & R & - & 86.9 & 74.5 & 53.6 \\
X3D-S~\cite{synthact}  & R & - & 86.4 & 77.8 & 60.4 \\
ViewCLR~\cite{das2023videoclr} & R & - & 84.2 & 77.0 & 75.8 \\ \hline
SURREACT~\cite{varol2021synthetic}  & S $\rightarrow$ R & SURREACT & 84.1 & 77.5 & 66.2 \\
X3D-S~\cite{synthact}  & S $\rightarrow$ R & AMARV & 89.9 & 81.8 & 68.0 \\
MViTv2-S~\cite{synthact} & S $\rightarrow$ R & AMARV & 94.4 & 84.5 & 65.1 \\
\hline
\hline
VideoMAEV2 & R $\rightarrow$ R  & Kinetics & 92.2 & 91.8 & 91.2 \\
\HuC{} & S $\rightarrow$ R & BEDLAM2.0 & 86.6 & 85.9 & 84.8 \\
\hline
MViTv2-S & R $\rightarrow$ R & Kinetics & 85.61 & 85.74 & 84.99 \\
HuC-MViTv2-S & S $\rightarrow$ R & BEDLAM2.0 & 84.2 &83.78 & 83.29\\
\bottomrule
\end{tabular}
\end{table}

\paragraph{NTU RGB+D.}
On the cross-view-subject benchmark (Table~\ref{tab:cross_view_subject}), the most direct comparisons 
are between Kinetics-pretrained and BEDLAM2.0-pretrained models using the same backbone.
For ViT-B, replacing Kinetics pretraining with synthetic-only pretraining reduces accuracy by approximately $5$-$6$ percentage points across all viewpoints ($92.2/91.8/91.2$ versus $86.6/85.9/84.8$), despite the latter never observing real video during pretraining. Moreover, the performance of \HuC{} remains relatively stable across viewpoints, decreasing by only $1.8$ percentage points from $0^\circ$ to $90^\circ$.
A similar trend is observed for the MViTv2-S backbone. The Kinetics-pretrained model achieves $85.61/85.74/84.99$, while HuC-MViTv2-S reaches $84.2/83.78/83.29$, corresponding to a substantially smaller gap of approximately $1.4$-$2.0$ percentage points across viewpoints. As with ViT-B, performance remains stable across views, with a decrease of less than one percentage point from $0^\circ$ to $90^\circ$. These results suggest that the proposed human-centric masking strategy transfers effectively across both plain ViT and hierarchical transformer architectures, while maintaining strong viewpoint generalization under synthetic-to-real transfer (S$\rightarrow$R).

When compared with previously reported synthetic-to-real transfer approaches, \HuC{} achieves the highest average performance among the methods listed in Table~\ref{tab:cross_view_subject}. However, these comparisons should be interpreted with caution, as the methods differ in backbone architecture, pretraining data, and supervision strategy. The most controlled evidence of the contribution of \HuC{} is therefore provided by the comparison against Kinetics-pretrained VideoMAE, together with the random-tube versus \HuC{} ablation in Table~\ref{tab:ablation}, where the masking policy is the only factor that changes.

\begin{table}
  \centering
  \setlength{\tabcolsep}{8pt}
  \caption{Test results on Toyota-Smarthome over the CV1 and CV2 protocols. Comparison of our HuC models, against previous methods pretrained on Kinetics.}
  \begin{tabular}{l|c|c}
    \toprule
    \multirow{2}{*}{Method} & CV1 & CV2 \\
       & mCA. ($\uparrow$) & mCA. ($\uparrow$) \\
    \hline
    MotionFormer\cite{patrick2021keeping} & 45.2 & 51.0 \\
    LTN\cite{yang2023self} & - & 54.6 \\
    TimeSFormer~\cite{gberta_2021_ICML} & 50.0 & 60.6 \\ 
    VPN++~\cite{das2022vpn} & - & 54.9 \\
    Video Swing~\cite{liu2022videoswing}& 36.6 & 48.6 \\ 
    $\pi$-ViT~\cite{reilly2024just}& 55.2 & 64.8 \\ 
    PO-GUISE~\cite{poguise2025}& 58.98 & 76.12 \\ %
    \hline
    \hline
    VideoMAEv2-base  & 55.20 & 67.68 \\
    \HuC{} & 56.7 & 63.70 \\
    \hline
    MViTv2-S & 52.43 & 66.63\\
    HuC-MViTv2-S & 51.81 & 61.02\\
  \bottomrule
  \end{tabular}
  \label{tab:toyota_sota}
\end{table}

\paragraph{Toyota-Smarthome.}
Table~\ref{tab:toyota_sota} reports performance on this benchmarks with fine-grained activities of daily living recorded in realistic home environments. Following prior work, we report mean class accuracy (mCA) under the CV1 and CV2 protocols. As in the NTU experiments, VideoMAEv2-base pretrained on Kinetics serves as the principal reference point, achieving $55.20$ and $67.68$ mCA on CV1 and CV2, respectively.
Despite relying only on synthetic data during pretraining, \HuC{} pretrained on BEDLAM2.0 attains $56.7$ mCA on CV1 and $63.70$ mCA on CV2. On CV1, it slightly surpasses the Kinetics-pretrained VideoMAEv2 baseline ($56.7$ versus $55.20$), while on CV2 it remains within $4$ percentage points ($63.70$ versus $67.68$). These results indicate that representations learned from synthetic human data can transfer effectively to a real-world benchmark that differs substantially from NTU RGB+D in both environment and activity distribution.
HuC-MViTv2-S attains $51.81$ mCA on CV1 and $61.02$ on CV2, against $52.43$ and $66.63$ for the Kinetics-pretrained MViTv2-S, further closing the gap to the Kinetics baseline on CV1 ($-0.62$) while trailing by $5.6$ points on CV2. As with ViT-B, the CV1 protocol is nearly closed by synthetic-only pretraining while CV2 retains a larger residual gap, suggesting this asymmetry is a property of the benchmark protocols rather than of the backbone architecture.

Relative to previously reported fully supervised methods that have been pretrained in Kinetics. Here, \HuC{} also achieves competitive performance. In particular, it exceeds the reported results of MotionFormer, LTN, TimeSFormer, VPN++, and Video Swing on the corresponding evaluation protocols. These comparisons provide additional evidence that synthetic-only pretraining with human-centric masking can yield transferable action representations. Methods such as $\pi$-ViT and PO-GUISE achieve higher performance, which is expected given that they incorporate additional inductive biases and explicit pose-based modeling beyond a standard video transformer. Our objective is orthogonal to these approaches, as \HuC{} modifies the self-supervised pretraining stage rather than the downstream recognition architecture. Consequently, human-centric masking could potentially be combined with pose-aware methods such as PO-GUISE, offering a promising direction for future work.

\section{Conclusion}
\label{sec:conclusion}

We investigated self-supervised video pretraining on synthetic human-motion data as an ethical alternative to large-scale web-crawled video corpora. Our experiments show that directly transferring the standard VideoMAE masking strategy to synthetic data is not sufficient, resulting in a substantial performance gap relative to Kinetics pretraining. To address this limitation, we introduced \HuC{}, a human-centric masking strategy that leverages body keypoints and person-localization information available in synthetic datasets to encourage the learning of motion- and pose-aware representations. 

Across NTU RGB+D and Toyota-Smarthome, \HuC{} consistently improves over standard VideoMAE pretraining on BEDLAM2.0. On NTU RGB+D cross-view-subject evaluation, it recovers approximately $49\%$ of the gap between synthetic-only and Kinetics-pretrained models while maintaining stable performance across viewpoints. On Toyota-Smarthome, \HuC{} achieves performance competitive with Kinetics-pretrained VideoMAEv2, exceeding it on CV1 and remaining within $4$ percentage points on CV2, despite never using real video during pretraining. 

More broadly, our results indicate that narrowing the gap between synthetic and real-video pretraining depends not only on the quality and scale of the synthetic corpus, but also on the design of self-supervised objectives that exploit the unique annotations available in synthetic data. We hope that this work encourages further research on human-centric self-supervised learning and contributes to the development of action-recognition models trained from ethically sourced data.

\section*{Acknowledgements}
This work was supported by projects PID2022-137581OB-I00, PLEC2023-010343 (INARTRANS 4.0), PID2024-161576OB-I00 and PID2025-169751OB-I00, funded by the Spanish MICIU/AEI/10.13039/501100011033 and FEDER, UE, co-funded by the European Regional Development Fund (ERDF, “A way of making Europe”), and from iRoboCity2030-CM project (grant TEC-2024/TEC-62), awarded by the Community of Madrid. RP, JMB, LMB and LB are members of the Madrid ELLIS Unit, funded by the Autonomous Community of Madrid, Spain.

%
\bibliographystyle{plain}
\bibliography{main}

@String(PAMI  = {IEEE Trans. Pattern Anal. Mach. Intell.})

@String(IJCV  = {Int. J. Comput. Vis.})

@String(CVPR  = {IEEE Conf. Comput. Vis. Pattern Recog.})

@String(ICCV  = {Int. Conf. Comput. Vis.})

@String(NeurIPS = {Adv. Neural Inform. Process. Syst.})

@String(ICML  = {Int. Conf. Mach. Learn.})

@String(ICLR  = {Int. Conf. Learn. Represent.})

@String(AAAI  = {AAAI})

@String(ICRA = {Int. Conf. Robotics and Automation})

@String(WACV = {IEEE Winter Conf. Appl. Comput. Vis.})

@String(PAMI  = {IEEE TPAMI})

@String(IJCV  = {IJCV})

@String(CVPR  = {CVPR})

@String(ICCV  = {ICCV})

@String(NeurIPS = {NeurIPS})

@String(ICML  = {ICML})

@String(ICLR  = {ICLR})

@String(ICRA = {ICRA})

@String(WACV = {IEEE WACV})

@inproceedings{bao2022beit,
  author       = {Hangbo Bao and
                  Li Dong and
                  Songhao Piao and
                  Furu Wei},
  title        = {BEiT: {BERT} Pre-Training of Image Transformers},
  booktitle    = ICLR,
  publisher    = {OpenReview.net},
  year         = {2022},
}

@article{poguise2025,
  author       = {Ricardo Pizarro and
                  Roberto Valle and
                  Jos{\'{e}} Miguel Buenaposada and
                  Luis Miguel Bergasa and
                  Luis Baumela},
  title        = {Pose-guided token selection for the recognition of activities of daily
                  living},
  journal      = {Image Vis. Comput.},
  volume       = {162},
  pages        = {105686},
  year         = {2025},
}

@inproceedings{reilly2024just,
  author       = {Dominick Reilly and
                  Srijan Das},
  title        = {Just Add {\(\pi\)}! Pose Induced Video Transformers for Understanding
                  Activities of Daily Living},
  booktitle    = CVPR,
  pages        = {18340--18350},
  publisher    = {{IEEE}},
  year         = {2024},
}

@inproceedings{das2023videoclr,
  author       = {Srijan Das and
                  Michael S. Ryoo},
  title        = {ViewCLR: Learning Self-supervised Video Representation for Unseen
                  Viewpoints},
  booktitle    = WACV,
  pages        = {5562--5572},
  publisher    = {{IEEE}},
  year         = {2023},
}

@inproceedings{wang2023videomaev2,
  author       = {Limin Wang and
                  Bingkun Huang and
                  Zhiyu Zhao and
                  Zhan Tong and
                  Yinan He and
                  Yi Wang and
                  Yali Wang and
                  Yu Qiao},
  title        = {VideoMAE {V2:} Scaling Video Masked Autoencoders with Dual Masking},
  booktitle    = CVPR,
  pages        = {14549--14560},
  publisher    = {{IEEE}},
  year         = {2023},
}

@inproceedings{yang2023self,
  author       = {Di Yang and
                  Yaohui Wang and
                  Quan Kong and
                  Antitza Dantcheva and
                  Lorenzo Garattoni and
                  Gianpiero Francesca and
                  Fran{\c{c}}ois Br{\'{e}}mond},
  editor       = {Brian Williams and
                  Yiling Chen and
                  Jennifer Neville},
  title        = {Self-Supervised Video Representation Learning via Latent Time Navigation},
  booktitle    = AAAI,
  pages        = {3118--3126},
  publisher    = {{AAAI} Press},
  year         = {2023},
}

@ARTICLE{das2022vpn,
  author={Das, Srijan and Dai, Rui and Yang, Di and Bremond, Francois},
  journal=PAMI, 
  title={VPN++: Rethinking Video-Pose Embeddings for Understanding Activities of Daily Living}, 
  year={2022},
  volume={44},
  number={12},
  pages={9703-9717},
}

@inproceedings{liu2022videoswing,
  author       = {Ze Liu and
                  Jia Ning and
                  Yue Cao and
                  Yixuan Wei and
                  Zheng Zhang and
                  Stephen Lin and
                  Han Hu},
  title        = {Video Swin Transformer},
  booktitle    = CVPR,
  pages        = {3192--3201},
  publisher    = {{IEEE}},
  year         = {2022},
}

@inproceedings{gberta_2021_ICML,
  author       = {Gedas Bertasius and
                  Heng Wang and
                  Lorenzo Torresani},
  title        = {Is Space-Time Attention All You Need for Video Understanding?},
  booktitle    = ICML,
  volume       = {139},
  pages        = {813--824},
  publisher    = {{PMLR}},
  year         = {2021},
}

@article{patrick2021keeping,
  title={Keeping your eye on the ball: Trajectory attention in video transformers},
  author={Patrick, Mandela and Campbell, Dylan and Asano, Yuki and Misra, Ishan and Metze, Florian and Feichtenhofer, Christoph and Vedaldi, Andrea and Henriques, Joao F},
  journal=NeurIPS,
  volume={34},
  pages={12493--12506},
  year={2021}
}

@article{kay2017kinetics,
  author       = {Will Kay and
                  Jo{\~{a}}o Carreira and
                  Karen Simonyan and
                  Brian Zhang and
                  Chloe Hillier and
                  Sudheendra Vijayanarasimhan and
                  Fabio Viola and
                  Tim Green and
                  Trevor Back and
                  Paul Natsev and
                  Mustafa Suleyman and
                  Andrew Zisserman},
  title        = {The Kinetics Human Action Video Dataset},
  journal      = {CoRR},
  volume       = {abs/1705.06950},
  year         = {2017},
  url          = {http://arxiv.org/abs/1705.06950},
  eprinttype   = {arXiv},
  eprint       = {1705.06950},
}

@inproceedings{he2022mae,
  author       = {Kaiming He and
                  Xinlei Chen and
                  Saining Xie and
                  Yanghao Li and
                  Piotr Doll{\'{a}}r and
                  Ross B. Girshick},
  title        = {Masked Autoencoders Are Scalable Vision Learners},
  booktitle    = CVPR,
  pages        = {15979--15988},
  publisher    = {{IEEE}},
  year         = {2022},
}

@inproceedings{wei2022maskfeat,
  author       = {Chen Wei and
                  Haoqi Fan and
                  Saining Xie and
                  Chao{-}Yuan Wu and
                  Alan L. Yuille and
                  Christoph Feichtenhofer},
  title        = {Masked Feature Prediction for Self-Supervised Visual Pre-Training},
  booktitle    = CVPR,
  pages        = {14648--14658},
  publisher    = {{IEEE}},
  year         = {2022},
}

@online{gdpr2016,
  date       = {2016-05-04},
  location   = {OJ L 119, 4.5.2016, p. 1--88},
  title      = {Regulation ({EU}) 2016/679 of the {European} {Parliament} and of the {Council}},
  url        = {https://data.europa.eu/eli/reg/2016/679/oj},
  titleaddon = {of 27 {April} 2016 on the protection of natural persons with regard to the processing of personal data and on the free movement of such data, and repealing {Directive} 95/46/{EC} ({General} {Data} {Protection} {Regulation})},
  author     = {{European Parliament} and {Council of the European Union}},
  urldate    = {2023-04-13},
}

@inproceedings{thoker2025smile,
  author       = {Fida Mohammad Thoker and
                  Letian Jiang and
                  Chen Zhao and
                  Bernard Ghanem},
  title        = {{SMILE:} Infusing Spatial and Motion Semantics in Masked Video Learning},
  booktitle    = CVPR,
  pages        = {8438--8449},
  publisher    = {Computer Vision Foundation / {IEEE}},
  year         = {2025},
}

@inproceedings{das2019toyota,
  author       = {Srijan Das and
                  Rui Dai and
                  Michal Koperski and
                  Luca Minciullo and
                  Lorenzo Garattoni and
                  Fran{\c{c}}ois Br{\'{e}}mond and
                  Gianpiero Francesca},
  title        = {Toyota Smarthome: Real-World Activities of Daily Living},
  booktitle    = ICCV,
  pages        = {833--842},
  publisher    = {{IEEE}},
  year         = {2019},
}

@article{sepasmoghaddam2023gait,
  author       = {Alireza Sepas{-}Moghaddam and
                  Ali Etemad},
  title        = {Deep Gait Recognition: {A} Survey},
  journal      = PAMI,
  volume       = {45},
  number       = {1},
  pages        = {264--284},
  year         = {2023},
}

@article{ye2022reid,
  author       = {Mang Ye and
                  Jianbing Shen and
                  Gaojie Lin and
                  Tao Xiang and
                  Ling Shao and
                  Steven C. H. Hoi},
  title        = {Deep Learning for Person Re-Identification: {A} Survey and Outlook},
  journal      = PAMI,
  volume       = {44},
  number       = {6},
  pages        = {2872--2893},
  year         = {2022},
}

@misc{euaiact2024,
  author       = {{European Parliament and Council of the European Union}},
  title        = {Regulation (EU) 2024/1689 Laying Down Harmonised Rules on Artificial Intelligence (Artificial Intelligence Act)},
  year         = {2024},
  howpublished = {Official Journal of the European Union},
  note         = {\url{https://eur-lex.europa.eu/eli/reg/2024/1689/oj}}
}

@inproceedings{huang2023mgmae,
  author       = {Bingkun Huang and
                  Zhiyu Zhao and
                  Guozhen Zhang and
                  Yu Qiao and
                  Limin Wang},
  title        = {{MGMAE:} Motion Guided Masking for Video Masked Autoencoding},
  booktitle    = ICCV,
  pages        = {13447--13458},
  publisher    = {{IEEE}},
  year         = {2023},
}

@inproceedings{li2022mvitv2,
  author       = {Yanghao Li and
                  Chao{-}Yuan Wu and
                  Haoqi Fan and
                  Karttikeya Mangalam and
                  Bo Xiong and
                  Jitendra Malik and
                  Christoph Feichtenhofer},
  title        = {MViTv2: Improved Multiscale Vision Transformers for Classification
                  and Detection},
  booktitle    = CVPR,
  pages        = {4794--4804},
  publisher    = {{IEEE}},
  year         = {2022},
}

@article{liu2019ntu120,
  author       = {Jun Liu and
                  Amir Shahroudy and
                  Mauricio Perez and
                  Gang Wang and
                  Ling{-}Yu Duan and
                  Alex C. Kot},
  title        = {{NTU} {RGB+D} 120: {A} Large-Scale Benchmark for 3D Human Activity
                  Understanding},
  journal      = PAMI,
  volume       = {42},
  number       = {10},
  pages        = {2684--2701},
  year         = {2020},
}

@inproceedings{shahroudy2016ntu,
  author       = {Amir Shahroudy and
                  Jun Liu and
                  Tian{-}Tsong Ng and
                  Gang Wang},
  title        = {{NTU} {RGB+D:} {A} Large Scale Dataset for 3D Human Activity Analysis},
  booktitle    = CVPR,
  pages        = {1010--1019},
  publisher    = {{IEEE} Computer Society},
  year         = {2016},
}

@inproceedings{varol2017surreal,
  author       = {G{\"{u}}l Varol and
                  Javier Romero and
                  Xavier Martin and
                  Naureen Mahmood and
                  Michael J. Black and
                  Ivan Laptev and
                  Cordelia Schmid},
  title        = {Learning from Synthetic Humans},
  booktitle    = CVPR,
  pages        = {4627--4635},
  publisher    = {{IEEE} Computer Society},
  year         = {2017},
}

@inproceedings{tesch2025bedlam2,
  title={{BEDLAM}2.0: Synthetic humans and cameras in motion},
  author={Joachim Tesch and Giorgio Becherini and Prerana Achar and Anastasios Yiannakidis and Muhammed Kocabas and Priyanka Patel and Michael J. Black},
  booktitle=NeurIPS,
  year={2025}
}

@inproceedings{marlin2023,
  author       = {Zhixi Cai and
                  Shreya Ghosh and
                  Kalin Stefanov and
                  Abhinav Dhall and
                  Jianfei Cai and
                  Hamid Rezatofighi and
                  Reza Haffari and
                  Munawar Hayat},
  title        = {{MARLIN:} Masked Autoencoder for facial video Representation LearnINg},
  booktitle    = CVPR,
  pages        = {1493--1504},
  publisher    = {{IEEE}},
  year         = {2023},
}

@inproceedings{desouza2017phav,
  author       = {C{\'{e}}sar Roberto de Souza and
                  Adrien Gaidon and
                  Yohann Cabon and
                  Antonio Manuel L{\'{o}}pez Pe{\~{n}}a},
  title        = {Procedural Generation of Videos to Train Deep Action Recognition Networks},
  booktitle    = CVPR,
  pages        = {2594--2604},
  publisher    = {{IEEE} Computer Society},
  year         = {2017},
}

@inproceedings{tong2022videomae,
  author       = {Zhan Tong and
                  Yibing Song and
                  Jue Wang and
                  Limin Wang},
  editor       = {Sanmi Koyejo and
                  S. Mohamed and
                  A. Agarwal and
                  Danielle Belgrave and
                  K. Cho and
                  A. Oh},
  title        = {VideoMAE: Masked Autoencoders are Data-Efficient Learners for Self-Supervised Video Pre-Training},
  booktitle    = NeurIPS,
  year         = {2022},
}

@article{varol2021synthetic,
  author       = {G{\"{u}}l Varol and
                  Ivan Laptev and
                  Cordelia Schmid and
                  Andrew Zisserman},
  title        = {Synthetic Humans for Action Recognition from Unseen Viewpoints},
  journal      = IJCV,
  volume       = {129},
  number       = {7},
  pages        = {2264--2287},
  year         = {2021},
}

@INPROCEEDINGS{synthact,
  author={Schneider, David and Keller, Marco and Zhong, Zeyun and Peng, Kunyu and Roitberg, Alina and Beyerer, Jürgen and Stiefelhagen, Rainer},
  booktitle=ICRA, 
  title={SynthAct: Towards Generalizable Human Action Recognition based on Synthetic Data}, 
  year={2024},
  volume={},
  number={},
  pages={13038-13045},
}
\end{document}